\pdfoutput=1
\documentclass[11pt,a4paper]{article}

\usepackage[a4paper,margin=1in]{geometry}
\usepackage{setspace}
\usepackage[T1]{fontenc}
\usepackage[utf8]{inputenc}
\usepackage{lmodern}

\usepackage{url}

\usepackage[numbers,sort&compress]{natbib}
\let\cite\citep
\usepackage[colorlinks=true,linkcolor=blue,citecolor=blue,urlcolor=blue]{hyperref}

\usepackage{amsmath,amssymb}
\usepackage{booktabs}
\usepackage{graphicx}
\usepackage{xcolor}
\usepackage{caption}

\usepackage{authblk}

\usepackage[framemethod=TikZ]{mdframed}
\definecolor{gray05}{gray}{.95}
\definecolor{steBoxLine}{rgb}{0.0, 0.0, 0.0}

\mdfdefinestyle{prompt}{%
    linecolor=steBoxLine,
    linewidth=3pt,
    leftmargin=0cm,rightmargin=0cm,
    roundcorner=5pt,
    topline=false,bottomline=false,rightline=false,
    backgroundcolor=gray05,
    skipabove=.5em,
    skipbelow=.5em,
    font=\sffamily\small
}

\newenvironment{prompt}[1]
  {\begin{mdframed}[style=prompt,frametitle={\sffamily\small\textbf{#1}}]}
  {\end{mdframed}}

\title{Machine-Learning-Enhanced Non-Invasive Testing for MASLD Fibrosis: Shallow-Deep Neural Networks Versus FIB-4, Tabular Foundation Models, and Large Language Models}

\author[1,2,3]{Athanasios Angelakis$^{*}$}
\author[4]{Gabriele De Vito$^{*}$}
\author[5]{Eleni-Myrto Trifylli}
\author[4]{Filomena Ferrucci}

\affil[1]{BioML Lab, RI CODE, UniBw, Munich, Germany}
\affil[2]{Department of Epidemiology and Data Science, Amsterdam UMC, Amsterdam, Netherlands}
\affil[3]{Alpha Indicium, Rijswijk, Netherlands}
\affil[4]{Department of Computer Science, University of Salerno, Salerno, Italy}
\affil[5]{GI-Liver Unit, 2nd Department of Internal Medicine, National and Kapodistrian University of Athens, General Hospital of Athens ``Hippocratio'', Athens, Greece}

\affil[*]{These authors contributed equally as co-first authors.}

\date{\vspace{-1em}
\small Correspondence: \texttt{athanasios.angelakis@unibw.de}; \texttt{gadevito@unisa.it}}

\begin{document}

\maketitle

\begin{abstract}
Advanced fibrosis is a major determinant of liver-related morbidity in metabolic dysfunction-associated steatotic liver disease (MASLD). FIB-4 is widely used as a first-line non-invasive test because it relies on routine clinical variables; however, its fixed formula may underuse diagnostic information contained in age, aspartate aminotransferase, alanine aminotransferase, and platelet count. We evaluated whether machine-learning-enhanced non-invasive testing (MLE-NIT) can improve advanced fibrosis detection while preserving this clinically accessible FIB-4 variable space.

We used three biopsy-confirmed MASLD cohorts from China, Malaysia, and India ($n=784$). The Chinese cohort was split into 486 training and 54 internal validation/tuning patients; final performance was reported only on the Malaysian and Indian external cohorts. Models used five variables: age, FIB-4, aspartate aminotransferase, platelet count, and alanine aminotransferase. We compared FIB-4 with a shallow-deep neural network (s-DNN), TabPFN, and \texttt{gpt-4o-2024-08-06} in zero-shot and fine-tuned settings.

FIB-4 achieved external ROC-AUCs of 0.75 and 0.60 in Malaysia and India, respectively. TabPFN achieved 0.69 and 0.66, fine-tuned GPT-4o achieved 0.75 and 0.63, and the s-DNN achieved 0.77 and 0.67, respectively. The s-DNN contained only 354 trainable parameters, compared with 7,244,554 parameters for TabPFN, yet provided a more balanced external operating profile. Calibration analysis showed s-DNN Brier scores of 0.18 and 0.22, and permutation importance identified AST and FIB-4 as dominant variables. Exploratory decision-curve analysis showed cohort-dependent clinical utility, favoring TabPFN in Malaysia and s-DNN in India.

Compact, domain-specific non-linear MLE-NITs may enhance FIB-4-based fibrosis assessment without increasing clinical data requirements, although local calibration, threshold selection, and prospective workflow validation remain necessary.

\end{abstract}


\section{Introduction}
\label{sec:introduction}

Metabolic dysfunction-associated steatotic liver disease (MASLD), formerly referred to as non-alcoholic fatty liver disease (NAFLD), is one of the most prevalent causes of chronic liver disease worldwide~\cite{rinella2023multisociety,younossi2023global}. Although hepatic steatosis is common, the development of fibrosis, and particularly advanced fibrosis, is the key determinant of liver-related outcomes~\cite{ng2023mortality}. Accurate identification of patients with advanced fibrosis is therefore essential for risk stratification, referral pathways, clinical monitoring, and therapeutic decision-making.

Liver biopsy remains the histological reference standard for fibrosis staging, but its invasiveness, cost, sampling variability, and limited suitability for repeated large-scale screening restrict its use in routine practice. Consequently, non-invasive tests (NITs) based on routine clinical and biochemical variables have become central to MASLD care~\cite{zoncape2024non}. Among these, FIB-4 is one of the most widely used tools because it requires only age, aspartate aminotransferase (AST), alanine aminotransferase (ALT), and platelet count (PLT) \cite{sterling2006fib4}. This simplicity makes FIB-4 attractive for population-level screening and primary-care triage.

However, FIB-4 also has important structural limitations. It is a fixed mathematical formula interpreted through predefined thresholds. As a result, it assumes that the relationship between its variables and fibrosis risk can be sufficiently represented by a single hand-crafted score~\cite{shaheen2026diabetes}. In clinically heterogeneous MASLD populations, this assumption may be too restrictive. The same laboratory variables may interact in non-linear and population-dependent ways, and a fixed threshold may fail to capture clinically meaningful patterns across external cohorts.

In medicine, NITs are diagnostic or prognostic tools that assess disease presence, severity, or risk without requiring tissue sampling or other invasive procedures. In hepatology, NITs include blood-based scores and biomarker panels, elastography-based measurements, imaging-based approaches, and composite diagnostic pathways \cite{easl2021nit,sterling2025aasldbloodnit,tacke2024easlmasld}. This motivates the concept of machine-learning-enhanced non-invasive tests (MLE-NITs), which we formally define in Section~\ref{sec:background} as machine learning models that enhance the decision function of an established NIT while preserving its original input variables and clinical workflow.

This work also builds on a longer methodological line of compact neural-network models for liver-disease classification. As early as 2018, Angelakis and colleagues investigated two- and three-hidden-layer DNNs for chronic liver disease and NAFLD classification using structured ultrasound-derived, elastographic, hemodynamic, and demographic variables \cite{angelakis2018acr_nafld_bmode_dnn,angelakis2018aium_cld_bmode_dnn,angelakis2018ecr_fibrosis_bmode_hemodynamic_dnn}. These early studies anticipated the present MLE-NIT rationale: clinically available measurements can be transformed into compact non-linear decision-support models rather than being interpreted only through isolated thresholds.

Previous work on the same biopsy-confirmed China, Malaysia, and India cohorts showed that machine learning models using FIB-4-related variables can outperform the original FIB-4 score \cite{angelakis2023fib4parameters}. Subsequent work further suggested that feature engineering and synthetic tabular data generation can improve models trained on the same FIB-4 variable family \cite{angelakis2024catboost}. More recently, shallow-deep neural networks (s-DNNs) have been used to combine multiple NITs and their underlying features into MLE-NIT frameworks for MASLD advanced fibrosis detection \cite{angelakis2025sdnnnits}.

In the present study, we extend this line of work by comparing three modern artificial intelligence paradigms against FIB-4 under a deliberately constrained feature setting: s-DNNs, tabular foundation models, and large language models (LLMs). The s-DNN and TabPFN models receive only five variables from the FIB-4 variable space: age, FIB-4, AST, PLT, and ALT. The LLM receives the same patient-level variables in a structured prompt, with the FIB-4 1.3 cut-off provided as contextual clinical information rather than as an additional measured variable. This design allows us to address a clinically important question: can state-of-the-art artificial intelligence models improve advanced fibrosis detection without requiring additional biomarkers, omics data, imaging features, or additional clinical variables?

Our central hypothesis is that compact, domain-specific non-linear models are better suited than general-purpose foundation models for low-dimensional structured clinical prediction. We show that the s-DNN provides the most balanced external operating profile among the evaluated models, supporting the use of MLE-NITs as practical extensions of conventional FIB-4-based fibrosis assessment. Because the input variables are the same routine variables already used by FIB-4, the aim is not biomarker discovery, but a workflow-preserving upgrade of the decision function applied to clinically familiar variables.

\paragraph{Contributions.} This work makes three methodological contributions:
\begin{enumerate}
    \item We formally define the \textbf{MLE-NIT} paradigm as a clinically constrained extension of established NITs, enforcing input closure, workflow invariance, and permissible closed-loop feature engineering.
    \item We introduce the \textbf{s-DNN} as an explicitly constrained compact architecture for low-dimensional structured clinical data, with depth and prime-width heuristics that aim to balance expressivity and sample efficiency.
    \item We show that, under strict FIB-4 variable-space constraints, a tiny task-specific s-DNN with only 354 trainable parameters can achieve comparable or stronger external performance than much larger general-purpose comparators such as TabPFN with 7,244,554 parameters, while providing a balanced operating profile and clinically plausible feature-importance patterns across real-world external clinical cohorts.
\end{enumerate}

\section{Background and Related Work}
\label{sec:background}

\subsection{Non-invasive tests and machine-learning-enhanced NITs}

Non-invasive tests (NITs) are used in clinical medicine to estimate disease presence, severity, prognosis, or treatment-related risk without requiring invasive sampling. In liver disease, the central role of NITs is to reduce reliance on liver biopsy while supporting fibrosis staging, prognostication, referral decisions, and longitudinal monitoring. Current liver NITs can be broadly grouped into three categories: blood-based tests, including routine laboratory scores and serum biomarker panels; tests that assess physical properties of liver tissue, including liver stiffness, attenuation, and related elastographic measurements; and imaging-based approaches that evaluate liver anatomy or tissue characteristics \cite{easl2021nit}. These tests are complementary rather than mutually exclusive, and modern MASLD pathways increasingly use stepwise strategies in which simple blood-based tests such as FIB-4 are applied first, followed by elastography or other second-line assessments when risk is indeterminate or elevated \cite{tacke2024easlmasld,sterling2025aasldbloodnit}.

Within this framework, we define machine-learning-enhanced non-invasive tests (MLE-NITs) as computational extensions of established NITs. An MLE-NIT preserves the input space of the original NIT or NIT pathway, but replaces or augments the fixed score, linear equation, or threshold-based decision rule with a data-driven model. Formally, if a conventional NIT maps a set of routinely available variables $X_{\mathrm{NIT}}$ to a clinical risk estimate or class label through a fixed function $f_{\mathrm{NIT}}(X_{\mathrm{NIT}})$, an MLE-NIT learns an alternative function $f_{\mathrm{ML}}(X_{\mathrm{NIT}})$ using machine learning, while keeping the same input variables. Therefore, an MLE-NIT may be based on gradient-boosted trees, (s-DNNs), 
or other supervised learning algorithms, and may include feature engineering or related model-development steps, but it does not introduce additional patient measurements beyond the original NIT.

This distinction is clinically important. Many artificial intelligence models improve prediction by adding more data modalities, such as omics, imaging, or free-text information. MLE-NITs pursue a different goal: they aim to improve the decision function applied to data already collected in routine care. In the case of FIB-4, this means that age, AST, ALT, and PLT remain the clinically required variables, while machine learning is used to model their non-linear interactions more flexibly than the original hand-crafted formula. Such models can therefore be embedded into existing clinical pathways without changing the laboratory request, patient examination, or first-line triage structure.

\begin{prompt}{Formal Definition: Machine-Learning-Enhanced Non-Invasive Test (MLE-NIT)}
Let $\mathcal{X}_{\mathrm{NIT}} \subset \mathbb{R}^{d}$ denote the set of routinely collected clinical and laboratory variables required by an established non-invasive test. An \textbf{MLE-NIT} is a data-driven decision function $f_{\mathrm{ML}}(\cdot)$ that operates exclusively on $\mathcal{X}_{\mathrm{NIT}}$ or on derived features $\mathcal{X}' = \mathcal{T}(\mathcal{X}_{\mathrm{NIT}})$, where $\mathcal{T}$ represents deterministic, closed-loop transformations, such as ratios, polynomial terms, or clinically motivated interactions, that introduce no external information. An MLE-NIT must satisfy three constraints:
\begin{enumerate}
    \item \textbf{Input Isolation:} No additional biomarkers, imaging, omics, free-text, or patient-reported data are incorporated.
    \item \textbf{Workflow Invariance:} The clinical data acquisition, laboratory ordering, and physician triage pathway remain identical to the original NIT. The MLE-NIT operates solely as a computational upgrade to the decision/interpretation layer.
    \item \textbf{Decision Enhancement:} The model replaces the fixed score or static threshold with a flexible, data-optimized decision boundary, potentially leveraging feature engineering that remains strictly confined to the original input space.
\end{enumerate}
Formally, the MLE-NIT computes $\hat{y} = f_{\mathrm{ML}}(\mathcal{T}(\mathcal{X}_{\mathrm{NIT}}))$, where $\mathcal{T}$ may be the identity map, corresponding to raw inputs, or a closed-loop feature-engineering function. This design preserves the cost, turnaround time, and clinical workflow of the conventional NIT while upgrading its diagnostic reasoning from rigid thresholding to adaptive, non-linear classification.
\end{prompt}

\subsection{FIB-4 as a clinically useful but structurally limited NIT}

FIB-4 is widely used as a first-line non-invasive test for liver fibrosis because it is inexpensive, simple, and based on routinely available variables: age, AST, ALT, and PLT \cite{sterling2006fib4}. Its clinical value lies in this simplicity. However, the same simplicity also imposes limitations. FIB-4 compresses several biological signals into a single fixed formula and then applies threshold-based interpretation. This makes it practical, but it may not be sufficiently flexible for heterogeneous MASLD populations.

The original dataset used in this study was introduced by Sang et al., who investigated biopsy-confirmed NAFLD patients from China, Malaysia, and India \cite{sang2021fibrosis}. Their logistic-regression model used routine medical markers and performed better than FIB-4 and the NAFLD fibrosis score for distinguishing early from advanced fibrosis. Importantly, this study demonstrated that even conventional statistical modeling of routine blood markers can improve upon standard fixed-score approaches. It also established an open, biopsy-confirmed, multi-cohort setting for evaluating advanced fibrosis models across external populations.

\subsection{From FIB-4 to MLE-NITs}

The present work builds directly on the idea that FIB-4 should not be viewed as the final possible use of its variables, but rather as a clinically validated input set for enhanced modelling. In MLE-NIT terminology, FIB-4 defines a low-cost and clinically accepted variable space, while machine learning provides an alternative decision function over that same space. Angelakis and Chen evaluated a CatBoost model using FIB-4-related variables on the same China, Malaysia, and India cohorts \cite{angelakis2023fib4parameters}. That study used age, sex, ALT, AST, PLT, AST/PLT, and FIB-4, combined with feature engineering, and showed that CatBoost outperformed FIB-4 on internal cross-validation and both external validation cohorts. This suggested that the FIB-4 variable space contains non-linear diagnostic information that is not fully exploited by the original FIB-4 equation.

A subsequent study extended the same principle using conditional tabular generative adversarial networks (CTGAN)-based synthetic tabular data generation, CatBoost, and feature engineering \cite{xu2019ctgan,prokhorenkova2018catboost,angelakis2024catboost}. Using the open China, Malaysia, and India MASLD cohorts, the study showed that a machine learning model based on FIB-4 parameters improved over standard FIB-4, and that synthetic-data augmentation further improved AUROC on the external Malaysian test set. This result is particularly relevant because it shows that the challenge is not only the choice of classifier, but also the ability to learn robust non-linear decision structures from limited and imbalanced clinical data.

Together, these studies support the broader concept of MLE-NITs: clinically simple NITs can be transformed from fixed-threshold tools into adaptive non-linear prediction models while preserving the same low-cost input variables.

\subsection{Shallow-deep neural networks as compact MLE-NITs}

s-DNNs are defined here as compact feed-forward neural networks designed for low-dimensional structured biomedical data. Architecturally, an s-DNN has either two or three hidden layers, and each hidden layer contains an odd prime number of nodes. The use of only two or three hidden layers distinguishes s-DNNs from deeper architectures, while the use of small prime-valued hidden-layer widths encourages compact, asymmetric architectures that avoid unnecessarily large or repetitive layer structures. In this work, the term ``shallow-deep'' therefore refers to a model family that is deeper than a conventional single-hidden-layer neural network, but deliberately much shallower and more constrained than modern large-scale deep learning architectures. The odd-prime width constraint is used here as a transparent architecture-selection heuristic for small asymmetric networks, not as a claim that primality is theoretically optimal or universally superior to other compact width choices.

\begin{prompt}{Formal Definition: Shallow-Deep Neural Network (s-DNN)}
A \textbf{s-DNN} is a compact feed-forward neural network explicitly constrained for low-dimensional structured biomedical prediction. Its architecture is defined by three structural priors:
\begin{enumerate}
    \item \textbf{Depth constraint:} Exactly $L \in \{2, 3\}$ hidden layers.
    \item \textbf{Width constraint:} Each hidden layer $\ell \in \{1, \dots, L\}$ contains $d^{(\ell)}$ neurons, where $d^{(\ell)} \in \mathbb{P}_{\mathrm{odd}}$, the set of odd prime numbers.
    \item \textbf{Complexity constraint:} The constrained depth-width space is intended to reduce model complexity and mitigate overfitting risk in small-sample clinical cohorts while retaining sufficient capacity for non-linear decision boundaries.
\end{enumerate}
Mathematically, the s-DNN computes:
\[
h^{(0)} = x, \quad h^{(\ell)} = \sigma\!\left(W^{(\ell)} h^{(\ell-1)} + b^{(\ell)}\right), \quad \ell=1,\dots,L
\]
\[
f_{\mathrm{s\text{-}DNN}}(x) =
\sigma_{\mathrm{out}}\!\left(W^{(L+1)} h^{(L)} + b^{(L+1)}\right),
\]
where $\sigma$ denotes the ReLU activation in the hidden layers, $\sigma_{\mathrm{out}}$ denotes the sigmoid output activation used for binary classification, and $W^{(\ell)} \in \mathbb{R}^{d^{(\ell)} \times d^{(\ell-1)}}$. In this study, we instantiate an s-DNN with $L=3$ and $(d^{(1)}, d^{(2)}, d^{(3)}) = (17, 5, 23)$, yielding 354 trainable parameters. This architecture is deliberately positioned between linear clinical scores and large-scale deep models: it captures clinically meaningful non-linear interactions among NIT-derived variables while remaining compact and suitable for deployment in resource-constrained settings. The prime-width rule is therefore treated as a reproducible compact-design heuristic rather than as an empirically optimized universal rule.
\end{prompt}

The design motivation behind s-DNNs is practical rather than purely architectural. Many clinical prediction problems involve small or moderate sample sizes, low-dimensional structured variables, and a need for models that can capture non-linear interactions without excessive complexity. In such settings, very large neural networks may be unnecessary or prone to overfitting, while purely linear scores or fixed-threshold rules may be too restrictive. s-DNNs are intended to occupy this intermediate space: they are compact enough for small tabular biomedical datasets, but flexible enough to learn non-linear combinations of clinically meaningful variables.

The theoretical basis for using feed-forward neural networks to approximate non-linear functions is well established \cite{cybenko1989approximation,hornik1989universal}, while modern deep learning has provided a general framework for learning hierarchical representations from data \cite{lecun2015deeplearning}. In the present context, however, the goal is not to build a large generic neural model. Instead, the goal is to use a compact and clinically constrained architecture that can learn non-linear interactions among NIT-derived variables while remaining compatible with small-sample biomedical settings.

This design philosophy is consistent with related compact-model work in medical imaging. ZACH-ViT, a compact Vision Transformer introduced for low-data medical imaging, was designed around regime-dependent inductive-bias alignment rather than universal benchmark dominance \cite{angelakis2026zachvit}. In that work, compactness, architectural simplicity, and task-regime alignment were treated as central design principles: ZACH-ViT removed positional embeddings and the dedicated class token, used global average pooling over patch representations, and achieved competitive few-shot performance with approximately 0.25M parameters when spatial ordering was weakly informative. Although ZACH-ViT addresses image data and s-DNNs address low-dimensional structured clinical variables, both follow the same methodological principle: the architecture should be deliberately constrained and aligned with the structure of the target biomedical problem, rather than made large or generic by default.

The use of compact DNNs in liver-disease decision-support work by Angelakis and colleagues dates back to 2018. In chronic liver disease, a three-hidden-layer DNN was applied to significant fibrosis binary classification using biopsy-validated patients and structured B-mode ultrasound morphologic and demographic data \cite{angelakis2018aium_cld_bmode_dnn}. A related study used a two-hidden-layer DNN for significant liver fibrosis classification using gender, B-mode morphologic measurements, and portal-vein hemodynamic information \cite{angelakis2018ecr_fibrosis_bmode_hemodynamic_dnn}. In NAFLD, a DNN was also used to classify steatosis using gender and B-mode ultrasound-derived speed-of-sound and echogenicity features \cite{angelakis2018acr_nafld_bmode_dnn}. These studies are important for the present manuscript because they show that the use of compact neural architectures for structured liver-disease classification predates the current foundation-model era.

In previous MASLD work, Angelakis introduced an MLE-NIT framework that combined FIB-4, NAFLD fibrosis score, APRI, AST/ALT, and their underlying variables using a compact three-layer DNN \cite{angelakis2025sdnnnits,angulo2007nfs,wai2003apri}. The model outperformed standalone NITs in both internal validation and external testing, supporting the idea that the clinical utility of NITs can be enhanced by learning their joint non-linear structure.

Related work also applied s-DNNs 
to biomarker-driven MASLD tasks. Trifylli et al. used open SomaScan proteomics data from biopsy-validated MASLD patients and showed that a minimal three-protein s-DNN model could achieve robust diagnostic performance for advanced liver fibrosis \cite{trifylli2025threeproteins}. This proteomics line was preceded by work using CatBoost, feature selection, and feature engineering on SomaScan proteomics data to identify a five-protein signature for advanced fibrosis in MASLD patients \cite{trifylli2024aasldproteomics}. Another study used s-DNNs 
with extracellular vesicle data to distinguish steatosis stages in MASLD patients \cite{trifylli2025sdnnev}, while the corresponding full peer-reviewed extracellular-vesicle study demonstrated the broader value of machine learning for MASLD steatosis staging using circulating plasma extracellular-vesicle features \cite{trifylli2025evwjg}. Although these studies used different biomarker spaces, they share a common methodological principle: compact non-linear models can extract clinically relevant diagnostic signal from small, structured biomedical datasets.

\subsection{Why compare against LLMs and tabular foundation models?}

Foundation models and LLMs are increasingly promoted as general-purpose artificial intelligence systems. Tabular foundation models, such as TabPFN, aim to provide strong performance on small tabular datasets without extensive task-specific training \cite{tabpfn}. LLMs, in contrast, are primarily text-based models, but they can be prompted or fine-tuned to process structured clinical records.

The present study uses these models as strong contemporary comparators. However, our hypothesis is not that larger and more general models are automatically better. Rather, we test whether general-purpose foundation models can match or exceed a compact domain-specific s-DNN when the task is a low-dimensional, structured, clinically grounded prediction problem. This comparison is important because it separates model popularity from model suitability. For MLE-NIT development, the relevant question is not whether a model is state-of-the-art in general, but whether it provides reliable, balanced, and externally generalizable diagnostic performance under realistic clinical constraints. The s-DNN design represents a deliberately constrained alternative: instead of using a large general-purpose model, it uses a compact two- or three-hidden-layer architecture with prime-valued hidden-layer widths to model non-linear interactions in small structured clinical datasets.

\section{Data}
\label{sec:datasets-processing}

The dataset used in this study was derived from a previously published multi-cohort investigation of biopsy-confirmed NAFLD patients from China, Malaysia, and India \cite{sang2021fibrosis}. Given the updated disease nomenclature, we refer to the study population as MASLD throughout this manuscript, while preserving the original dataset definition and biopsy-based outcome labels. The original study included three independent cohorts comprising a total of 784 patients with hepatic fibrosis. The discovery cohort consisted of 540 patients recruited at Zhongshan Hospital, Fudan University, Shanghai, China. Two independent validation cohorts were obtained from existing studies conducted at the University of Malaya Medical Center and included 147 and 97 patients, respectively.

All included patients had NAFLD/MASLD confirmed by liver biopsy. In the Chinese discovery cohort, liver biopsy specimens were obtained from patients who met diagnostic criteria for non-alcoholic fatty liver disease or non-alcoholic steatohepatitis and underwent liver biopsy. Patients were excluded if they had a history of cancer, alcoholic intemperance, or other causes of chronic liver disease. For the two validation cohorts, the original publication reports that they were collected from anonymous datasets of existing studies conducted at the University of Malaya Medical Center.

Liver biopsy was used as the clinical reference standard for fibrosis staging in all three cohorts. In the discovery cohort, ultrasound-guided liver biopsies were performed using 1.6-mm-diameter needles. In the validation cohorts, percutaneous needle biopsies were performed by experienced operators using an 18-G Temno II semi-automatic biopsy needle. Liver tissue samples from each cohort were evaluated by an experienced pathologist blinded to the study design. Histological assessment followed the NAFLD activity score, including steatosis, lobular inflammation, hepatocellular ballooning, and fibrosis staging from 0 to 4.

For model development and evaluation, fibrosis was dichotomized into early fibrosis, defined as stages F0--F2, and advanced fibrosis, defined as stages F3--F4. The original Chinese cohort was randomly split into a training subset and an internal validation subset. Specifically, 486 of 540 patients (90.0\%) were used for model training or fitting, while 54 of 540 patients (10.0\%) were held out as an internal validation subset. The internal validation subset was used only for model selection and tuning where applicable, particularly for the LLM and s-DNN workflows. No Chinese internal-validation performance is reported as a final result. Final model assessment was performed on two external validation cohorts: the Malaysian cohort of 147 patients and the Indian cohort of 97 patients. In the present analysis, the prevalence of advanced fibrosis was 27.6\% in the Chinese cohort, 21.08\% in the Malaysian validation cohort, and 32.98\% in the Indian validation cohort.

All tabular machine learning models were trained and evaluated using five raw structured clinical features from the FIB-4 variable space: age, FIB-4, AST, PLT, and ALT. Although the formal MLE-NIT definition permits closed-loop feature engineering within the original NIT variable space, no engineered features were used for the s-DNN or TabPFN experiments reported here. The LLM prompt used the same five patient-level variables and additionally stated the FIB-4 1.3 cut-off as contextual clinical information. In line with the MLE-NIT definition, no additional biomarkers, imaging variables, omics features, free-text clinical information, or additional laboratory variables were introduced.

\section{Method}
\label{sec:method}

\subsection{Study design}

We formulated advanced fibrosis detection as a binary classification task. The positive class corresponded to advanced fibrosis, defined as histological stages F3--F4, and the negative class corresponded to early or no advanced fibrosis, defined as stages F0--F2. The Chinese cohort was randomly split into a 486-patient training subset and a 54-patient internal validation/tuning subset. The same Chinese training subset was used for all model families. The internal validation/tuning subset was used only for model selection or tuning where applicable, and all reported model performance was assessed on the Malaysian and Indian external validation cohorts.

The study was designed as a controlled comparison between conventional FIB-4 and MLE-NITs. The s-DNN and TabPFN models were constrained to the same five variables from the FIB-4 variable space: age, FIB-4, AST, PLT, and ALT. The LLM was evaluated using a structured representation of the same variables, with the FIB-4 1.3 cut-off included only as clinical context in the prompt. This restriction was intentional and is central to the MLE-NIT design. It ensures that any performance improvement comes from replacing the fixed FIB-4 decision function with a non-linear or prompt-based artificial intelligence decision function, rather than from adding biomarkers, imaging features, omics data, or other variables that would alter the clinical pathway.

We compared four model families: conventional FIB-4, a s-DNN, 
the tabular foundation model TabPFN, and the LLM \texttt{gpt-4o-2024-08-06} evaluated in zero-shot and fine-tuned settings. Model performance was evaluated using specificity, sensitivity, ROC-AUC computed at the operating classification threshold, and F1-score. For FIB-4, thresholded ROC-AUC was computed using the 1.3 clinical cut-off. For the machine-learning models, thresholded ROC-AUC was computed using the fixed 0.50 operating point, which is the standard decision boundary in binary classification where $P(\text{class}=1) > 0.5$ leads to predicting the positive class \cite{powers2011evaluation}. This 0.50 threshold was not selected or optimized using the Chinese internal validation/tuning subset, nor was it optimized on the external validation cohorts. The internal validation/tuning subset was used only for model selection or tuning where applicable; threshold optimization was not used to generate the reported external validation metrics.

Where continuous risk scores were available, probability-based ROC-AUC was additionally computed as a secondary discrimination analysis. Specifically, continuous FIB-4 values were used to compute discrimination independently of the clinical threshold, TabPFN probabilities were obtained using \texttt{predict\_proba}, and s-DNN probabilities were obtained from the final sigmoid output of the trained neural network. Bootstrap 95\% confidence intervals for external validation ROC-AUCs were estimated using 2,000 bootstrap resamples within each external cohort \cite{efron1993bootstrap,carpenter2000bootstrap}.

To characterize the baseline signal contained in the FIB-4 variable space, we performed univariable association analyses between each input feature and advanced fibrosis in the Chinese training subset. Continuous variables were compared using Welch's test and the Mann--Whitney U test. Linear and monotonic associations were assessed using point-biserial correlation and Spearman correlation, respectively. Standardized univariable logistic regression was used to estimate the odds ratio associated with a one-standard-deviation increase in each feature. These analyses were descriptive and were not used for feature selection.

\subsection{Univariable association and clinical utility analyses}
\label{sec:association-dca}

To contextualize the FIB-4 variable space, the association between each of the five input variables and the binary advanced-fibrosis outcome was assessed in the Chinese training subset. This analysis was descriptive and was not used for feature selection. For each continuous variable, we planned both parametric and non-parametric comparisons between fibrosis classes, including Welch's two-sample test and the Mann--Whitney U test. We also estimated \newline point-biserial/Pearson and Spearman correlations with the binary outcome, and univariable logistic-regression effects after standardization. The purpose of this analysis was to show that the MLE-NIT framework starts from clinically established variables with measurable univariable association to the outcome, but then evaluates whether non-linear modelling can provide a better workflow-preserving decision layer than fixed-score thresholding.

Clinical utility was assessed using exploratory decision-curve analysis (DCA) on the Malaysian and Indian external validation cohorts. Net benefit was computed across threshold probabilities from 0.05 to 0.50 and compared with treat-all and treat-none strategies \cite{vickers2006decision}. Because DCA requires continuous risk estimates, FIB-4 was represented by a logistic calibration model fitted on the Chinese training subset using FIB-4 as the sole predictor. This calibrated FIB-4 risk model was used only for DCA; the 1.3 threshold remained the classification threshold used for the FIB-4 row in Table~\ref{tab:external_results}. TabPFN probabilities were obtained using \texttt{predict\_proba}, and s-DNN probabilities were obtained from the final sigmoid output of the trained neural network. LLM-based approaches were not included in the main DCA because only classification outputs, rather than calibrated continuous risk estimates, were available. DCA was treated as exploratory because the external cohorts are modest in size and because threshold preferences may differ across clinical pathways.

\subsection{FIB-4 baseline}

FIB-4 was used as the clinical baseline comparator. It was calculated from age, AST, ALT, and PLT according to the original formula \cite{sterling2006fib4}. The continuous FIB-4 score was provided as an input variable to the artificial intelligence models together with its underlying variables, namely age, AST, ALT, and PLT.

\subsection{Large language model}
\label{sec:llm}

This section describes the approaches used to adapt an LLM for advanced fibrosis detection. We implemented two adaptation strategies: zero-shot learning, which leverages the model's pretrained knowledge without task-specific parameter updates, and supervised fine-tuning, which adapts the model to the binary fibrosis classification task.

\subsubsection{LLM selection}

The model \texttt{gpt-4o-2024-08-06} was selected for the experiments and accessed via the OpenAI API. The decision was based on three considerations. First, the model supports supervised fine-tuning, which is necessary for the paired zero-shot versus fine-tuning design. Second, using a dated model snapshot reduces the risk of model drift across releases. Third, GPT-4o has been evaluated in several recent clinical and biomedical benchmarking studies, and the authors have previously applied LLMs to clinical decision-support and drug-interaction prediction tasks, supporting its use as a representative general-purpose LLM comparator in this work \cite{openai_gpt4o,openai_finetuning,de2025heliot,de2026llms}.

The temperature was set to 0.0 at inference time to reduce output variability. We ran the classification five times and calculated the mean and standard deviation to account for LLM non-determinism.

\subsubsection{Zero-shot approach}
\label{sec:zero-shot}

The first adaptation strategy assessed the LLM in a zero-shot setting, with the aim of quantifying how much capability the model already has from pretraining alone for advanced fibrosis detection, without task-specific supervision or in-context demonstrations. This setting represents the most practical situation in which a clinician could query a general-purpose LLM without AI-engineering infrastructure, while also establishing a baseline for comparison with the fine-tuned counterpart described in Section~\ref{sec:fine-tuning}.

The task was cast in a role-conditioned form, in which the system message assigns the model the persona of a hepatologist (see Figure~\ref{prompt:zero-shot-sys}). The user message lists the five FIB-4 variable-space features in a field-value layout and additionally provides the FIB-4 1.3 cut-off as contextual clinical information (see Figure~\ref{prompt:zero-shot-user}).

\begin{prompt}{System prompt}
You are a knowledgeable and precise hepatologist. Given the patient's information I'll provide, classify whether the patient has `advanced fibrosis' or `no advanced fibrosis'. Answer only with the classification.
\end{prompt}
\vspace{-15pt}
\begin{figure}[!htbp]
\caption{System prompt used in the zero-shot setting.}
\label{prompt:zero-shot-sys}
\end{figure}

\begin{prompt}{User prompt: Zero-shot setting}
Patient: A Metabolic Dysfunction-Associated Steatotic Liver Disease (MASLD) patient\\
AGE: \{age\} years old\\
Platelet count (PLT): \{PLT\}\\
AST level: \{AST\}\\
ALT level: \{ALT\}\\
FIB-4 index: \{FIB4\}\\
FIB-4 index cut-off: 1.3\\[6pt]
Do not include any additional explanation, reasoning, or chain-of-thought. Your final answer must be only ``advanced fibrosis'' or ``no advanced fibrosis''.\\[4pt]
CLASSIFICATION:
\end{prompt}
\vspace{-15pt}
\begin{figure}[h]
\caption{User prompt template instantiated per patient. Placeholders in braces are filled with the tabular values of the corresponding row. The FIB-4 1.3 cut-off was provided as contextual clinical information for the LLM only.}
\label{prompt:zero-shot-user}
\end{figure}

Despite the explicit output constraints, an LLM operating in zero-shot mode is not guaranteed to comply. To preserve predictions without manual intervention, we wrapped the model call in a deterministic parsing cascade. First, the raw output was normalized by lowercasing and removing leading and trailing whitespace. Second, exact matches to ``advanced fibrosis'' and ``no advanced fibrosis'' were mapped to the positive and negative class, respectively. Third, if the response contained one of the two allowed labels as a substring, the corresponding label was extracted. Fourth, outputs that still could not be mapped unambiguously were flagged as invalid and excluded from metric computation.

\subsubsection{Fine-tuning approach}
\label{sec:fine-tuning}

The second adaptation strategy used supervised fine-tuning. The motivation was to verify whether the residual gap between the zero-shot LLM and the tabular baselines could be reduced by exposing the model to labeled examples drawn from the same clinical distribution as the training cohort, while keeping the input representation, role prompt, and output vocabulary identical to those of Section~\ref{sec:zero-shot}. This design isolates the contribution of parameter adaptation from prompt-formatting confounders.

The full Chinese training subset was serialized into conversational examples for supervised fine-tuning, while the internal validation/tuning subset was reserved for model selection where applicable. Each patient record was serialized into a single conversational example following the OpenAI chat-completion schema. Every example contained three messages: (i) the same system message used in the zero-shot setting, which establishes the hepatologist persona and binary output constraint; (ii) a user message instantiated from the structured prompt template of Figure~\ref{prompt:zero-shot-user}; and (iii) an assistant message containing the ground-truth label, encoded verbatim as either \texttt{advanced fibrosis} or \texttt{no advanced fibrosis}. The deliberate alignment between the zero-shot and fine-tuning formats enabled a clean before-and-after comparison.

Fine-tuning was carried out through the OpenAI fine-tuning API on the \texttt{gpt-4o-2024-08-06} backbone. No extensive hyperparameter search was performed, in order to avoid overfitting and to maintain a reproducible protocol. The fine-tuned checkpoint was queried at inference time with the same decoding configuration used in the zero-shot setting, and its outputs were processed by the same response decoder described above.

\subsection{Shallow-deep neural network as an MLE-NIT}
\label{sec:sdnn}

The s-DNN was implemented as the main MLE-NIT model. Following the definition introduced in Section~\ref{sec:background}, the s-DNN was designed as a compact feed-forward neural network with two or three hidden layers and an odd prime number of nodes in each hidden layer. The purpose of this design is to provide a deliberately constrained non-linear architecture for low-dimensional structured biomedical data, avoiding both the rigidity of fixed clinical scores and the unnecessary complexity of large deep learning models.

In this study, the architecture consisted of three hidden layers with 17, 5, and 23 nodes, respectively. Each hidden-layer width is an odd prime number, consistent with the s-DNN design principle. Rectified linear unit activation was used in the hidden layers, and the model was optimized using Adam. Training was performed for 100 epochs. The model received five raw input variables: age, FIB-4, AST, PLT, and ALT. No engineered variables were used in the s-DNN experiments reported in this manuscript.

Class imbalance was handled during training using a cost-sensitive learning strategy based on class weighting, without oversampling or synthetic data generation. The Chinese internal validation subset was used only for model selection/tuning and is not reported as a performance result. Model performance was summarized on the external validation cohorts using thresholded ROC-AUC at the fixed 0.50 classification threshold, accuracy, sensitivity, specificity, and F1-score.

For external model diagnostics, the trained s-DNN was additionally assessed using calibration and explainability analyses on the Malaysian and Indian external validation cohorts. Calibration was evaluated using reliability diagrams, Brier scores, and expected calibration error (ECE) \cite{steyerberg2010assessing,guo2017calibration}, while retaining the 0.50 threshold for classification metrics. Model-level explainability was assessed using permutation feature importance, defined as the decrease in ROC-AUC at the 0.50 classification threshold after repeated random permutation of each input variable \cite{breiman2001random,altmann2010permutation}. This analysis was used to evaluate whether the s-DNN decision function relied on clinically plausible components of the FIB-4 variable space.

The purpose of this model was to test whether a compact non-linear architecture can improve advanced fibrosis detection while using only the same clinically accessible variables that support FIB-4-based assessment. This design follows the earlier compact-DNN liver-disease classification line, where two- and three-hidden-layer DNNs were used with structured ultrasound-derived, elastographic, hemodynamic, and demographic variables for fibrosis and steatosis classification \cite{angelakis2018aium_cld_bmode_dnn,angelakis2018ecr_fibrosis_bmode_hemodynamic_dnn,angelakis2018acr_nafld_bmode_dnn}.

This design also follows the MLE-NIT rationale introduced in previous work: rather than replacing NITs with complex or expensive tests, machine learning is used to learn non-linear combinations of NIT-derived variables. The s-DNN is therefore not proposed as a black-box alternative to clinical testing, but as a machine-learning-enhanced form of routine non-invasive fibrosis assessment.

\subsection{Tabular foundation model}
\label{sec:tabpfn}

TabPFN was evaluated as a representative tabular foundation model \cite{tabpfn}. Unlike conventional supervised models that are trained from scratch for a specific dataset, TabPFN uses prior-data fitting to enable rapid inference on small tabular classification problems. This makes it an attractive candidate for biomedical settings where labeled cohorts are often small and external validation is essential.

In this study, TabPFN received the same five input variables as the s-DNN: age, FIB-4, AST, PLT, and ALT. TabPFN was fitted using the same Chinese training subset used for all other model families and was evaluated on the Malaysian and Indian external validation cohorts using specificity, sensitivity, ROC-AUC at the 0.50 classification threshold, and F1-score.

\subsection{Model-complexity and computational-environment reporting}
\label{sec:model_complexity}

To contextualize the comparison between the compact s-DNN and the pretrained tabular foundation model, we additionally quantified model complexity. For the s-DNN, we counted the number of trainable parameters directly from the PyTorch model and measured the serialized PyTorch \texttt{state\_dict} size. For TabPFN, we reported the number of parameters in the largest detected internal PyTorch module and the serialized fitted-classifier object size. This comparison was used only to characterize the difference in model scale and deployment footprint, not as an additional performance metric.

All experiments were performed using Python 3.10.16 on Linux. The main software versions were NumPy 1.26.4, pandas 2.3.2, scikit-learn 1.7.2, SciPy 1.15.3, PyTorch 2.3.1, TabPFN 2.0.6, matplotlib 3.10.5, and joblib 1.5.2. CUDA was available through PyTorch CUDA 12.1 with cuDNN 8902 on one NVIDIA GeForce RTX 3060 GPU.

\section{Results and Discussion}
\label{sec:results}

\subsection{FIB-4 performance confirms the need for enhanced NIT modelling}

FIB-4 showed variable external validation performance across cohorts. In the Malaysian validation cohort, FIB-4 achieved a thresholded ROC-AUC of 0.75, specificity of 0.79, sensitivity of 0.71, and F1-score of 0.57. In the Indian validation cohort, performance decreased to a thresholded ROC-AUC of 0.60, specificity of 0.66, sensitivity of 0.53, and F1-score of 0.48.

This variability is clinically important. FIB-4 remains useful as a simple first-line NIT, but these results show that FIB-4 alone is not sufficiently robust across external cohorts. The drop in AUROC and F1-score in the Indian cohort suggests that a fixed score and threshold may not generalize well when cohort characteristics differ. Therefore, the question is not whether FIB-4 should be abandoned, but whether it should be enhanced by non-linear modelling.

\subsection{Univariable association of FIB-4 variable-space features with advanced fibrosis}

Before evaluating the machine-learning models, we assessed the univariable relationship between each FIB-4 variable-space feature and advanced fibrosis in the Chinese development cohort. This analysis was performed to clarify the baseline diagnostic information contained in the variables used by the MLE-NIT models and to determine whether the model inputs were individually associated with the histological endpoint.

\begin{table}[h]
\centering
\scriptsize
\caption{Univariable association between FIB-4 variable-space features and advanced fibrosis in the Chinese development cohort. Continuous variables are summarized as mean $\pm$ standard deviation. Standardized odds ratios correspond to one-standard-deviation increase in each feature.}
\label{tab:univariable_associations}
\setlength{\tabcolsep}{3pt}
\begin{tabular}{lcccccc}
\toprule
Feature & Early mean$\pm$SD & Advanced mean$\pm$SD & M--W $p$ & Spearman $\rho$ & Std. OR & 95\% CI \\
\midrule
Age   & 44.29$\pm$13.51  & 53.04$\pm$11.28  & $6.72{\times}10^{-11}$ & 0.30  & 2.07 & 1.64--2.61 \\
FIB-4 & 1.04$\pm$0.58    & 1.88$\pm$1.11    & $8.42{\times}10^{-24}$ & 0.46  & 3.39 & 2.52--4.55 \\
AST   & 43.28$\pm$24.75  & 55.39$\pm$26.25  & $1.88{\times}10^{-8}$  & 0.26  & 1.56 & 1.28--1.90 \\
PLT   & 235.82$\pm$61.51 & 202.16$\pm$57.67 & $2.22{\times}10^{-7}$  & -0.24 & 0.53 & 0.41--0.67 \\
ALT   & 75.10$\pm$49.55  & 78.14$\pm$47.90  & 0.449                  & 0.03  & 1.06 & 0.87--1.29 \\
\bottomrule
\end{tabular}
\end{table}

Age, FIB-4, AST, and PLT were significantly associated with advanced fibrosis. FIB-4 showed the strongest univariable association, with a standardized odds ratio of 3.39 and a Spearman correlation of 0.46. Age and AST were positively associated with advanced fibrosis, whereas PLT was inversely associated, consistent with the biological direction encoded in the original FIB-4 formula. ALT was not significantly associated with advanced fibrosis in this cohort. These results show that the MLE-NIT models operate on variables that already contain clinically meaningful signal, but that this signal is partly constrained by the limited information content of the original FIB-4 variable space.

This finding is important for interpreting the MLE-NIT framework. The aim of MLE-NIT is not to create new biomarkers or to replace the clinical pathway with additional measurements, but to improve the decision function applied to the same variables that clinicians already use. Therefore, the expected performance gain is naturally bounded by the information contained in age, AST, ALT, PLT, and FIB-4. Within this constraint, machine learning can model non-linear interactions and alternative weighting schemes, but it cannot overcome the intrinsic limitations of the underlying low-dimensional clinical feature space.

\subsection{Brief feature-ablation analysis in the Chinese training subset}

As an exploratory robustness analysis, we performed a brief leave-one-feature-out ablation using stratified 10-fold cross-validation within the Chinese training subset. This analysis was not used for model selection, but was included to evaluate whether the models depended strongly on individual variables from the FIB-4 feature space, following standard sensitivity-analysis methodology for feature importance assessment.

For TabPFN, removing AST produced the largest decrease in thresholded ROC-AUC, from 0.70 to 0.67, corresponding to a drop of 0.04. Removing age produced a smaller decrease of 0.01, whereas removing ALT had minimal effect. Interestingly, removing FIB-4 or PLT did not reduce TabPFN performance in this internal cross-validation analysis, suggesting partial redundancy among the FIB-4-derived variables and their underlying components.

For the s-DNN, leave-one-feature-out ablation showed only small changes in thresholded ROC-AUC. Removing AST decreased thresholded ROC-AUC from 0.75 to 0.74, whereas removal of age, ALT, PLT, or FIB-4 produced negligible or slightly positive changes. These small ablation effects suggest that the compact neural model distributes information across correlated FIB-4 variable-space features rather than relying exclusively on a single predictor. This is consistent with the purpose of MLE-NITs: to learn an alternative decision function over a clinically constrained and partly redundant variable space, rather than to identify new independent biomarkers.

\subsection{External validation across Malaysian and Indian cohorts}

External validation was performed on the Malaysian and Indian cohorts only. Table~\ref{tab:external_results} summarizes the available cohort-level results for FIB-4, TabPFN, s-DNN, and LLM-based approaches. All values are rounded to two decimals. ROC-AUC is reported at the 0.50 classification threshold for machine-learning models and at the 1.3 threshold for FIB-4. The Chinese internal validation subset was used for tuning/model selection and is therefore not reported as a performance result. Bootstrap 95\% confidence intervals for external validation ROC-AUCs were estimated using 2,000 bootstrap resamples within each external cohort \cite{efron1993bootstrap,carpenter2000bootstrap}. For FIB-4, the thresholded analysis used the 1.3 clinical cut-off; for TabPFN and s-DNN, the thresholded analysis used the fixed 0.50 operating point. Probability-based AUCs are reported separately where continuous risk scores were available.

\begin{table}[h]
\centering
\caption{External validation performance across the Malaysian and Indian cohorts. ROC-AUC is computed at the 0.50 classification threshold for machine-learning models and at the 1.3 threshold for FIB-4. Bootstrap 95\% confidence intervals are shown for thresholded ROC-AUCs where available. A dash indicates that the metric was not available from the corresponding experiment.}
\label{tab:external_results}

\setlength{\tabcolsep}{3pt} 
\renewcommand{\arraystretch}{1.0} 

\begin{tabular}{llcccc}
\toprule
Cohort & Model & Specificity & Sensitivity & ROC-AUC & F1-score \\
\midrule
Malaysia & FIB-4 & 0.79 & 0.71 & 0.75 (0.66--0.83) & 0.57 \\
Malaysia & TabPFN & 0.92 & 0.45 & 0.69 (0.60--0.78) & 0.52 \\
Malaysia & s-DNN & 0.74 & 0.81 & 0.77 (0.69--0.85) & 0.58 \\
Malaysia & Zero-shot LLM & 0.65 $\pm$ 0.02 & 0.81 $\pm$ 0.02 & 0.73 (0.71--0.75) & 0.52 $\pm$ 0.02 \\
Malaysia & Fine-tuned LLM & 0.75 $\pm$ 0.01 & 0.75 $\pm$ 0.01 & 0.75 (0.74--0.76) & 0.56 $\pm$ 0.01 \\
\midrule
India & FIB-4 & 0.66 & 0.53 & 0.60 (0.48--0.69) & 0.48 \\
India & TabPFN & 0.85 & 0.47 & 0.66 (0.56--0.75) & 0.53 \\
India & s-DNN & 0.74 & 0.59 & 0.67 (0.56--0.76) & 0.56 \\
India & Zero-shot LLM & 0.56 $\pm$ 0.01 & 0.58 $\pm$ 0.01 & 0.57 (0.56--0.58) & 0.47 $\pm$ 0.01 \\
India & Fine-tuned LLM & 0.76 $\pm$ 0.03 & 0.50 $\pm$ 0.00 & 0.63 (0.61--0.64) & 0.50 $\pm$ 0.01 \\
\bottomrule
\end{tabular}
\vspace{2pt}

{\footnotesize $\pm$ indicates the standard deviation across five repeated LLM runs. Values in parentheses denote bootstrap 95\% confidence intervals for FIB-4, TabPFN, and s-DNN, and the observed range across repeated LLM runs for LLM-based models.}
\end{table}


\begin{table}[h]
\centering
\caption{Model-complexity comparison between the task-specific s-DNN and TabPFN. The s-DNN size corresponds to the serialized PyTorch \texttt{state\_dict}. The TabPFN size corresponds to the pickled fitted \texttt{TabPFNClassifier} object.}
\label{tab:model_complexity}
\begin{tabular}{lccc}
\toprule
Model & Parameters & Trainable parameters & Serialized size (MB) \\
\midrule
s-DNN & 354 & 354 & 0.0045 \\
TabPFN & 7,244,554 & 7,244,554 & 28.0648 \\
\bottomrule
\end{tabular}
\end{table}

The difference in model scale was substantial. TabPFN contained approximately 20,465 times more parameters than the s-DNN and its serialized fitted object was approximately 6,172 times larger than the s-DNN serialized \texttt{state\_dict}. Thus, the s-DNN represents an extremely compact task-specific MLE-NIT, whereas TabPFN represents a much larger pretrained foundation-model comparator. This scale difference is important for clinical deployment, because compact models are easier to audit, store, version, and integrate into lightweight clinical decision-support systems.

Because DCA and calibration analyses require continuous risk estimates, we also computed probability-based ROC-AUCs where continuous model scores were available. In Malaysia, probability-based ROC-AUCs were 0.85 (95\% CI: 0.77--0.92) for calibrated FIB-4, 0.81 (95\% CI: 0.72--0.89) for TabPFN, and 0.83 (95\% CI: 0.74--0.90) for s-DNN. In India, the corresponding probability-based ROC-AUCs were 0.63 (95\% CI: 0.49--0.74), 0.69 (95\% CI: 0.57--0.80), and 0.70 (95\% CI: 0.59--0.80), respectively. These probability-based results confirm that the relative ranking of models depends on whether the evaluation emphasizes continuous discrimination, fixed-threshold operating performance, or net clinical benefit.

FIB-4 showed variable external validation performance. In the Malaysian cohort, it achieved a ROC-AUC of 0.75, specificity of 0.79, sensitivity of 0.71, and F1-score of 0.57. In the Indian cohort, performance decreased to a thresholded ROC-AUC of 0.60, specificity of 0.66, sensitivity of 0.53, and F1-score of 0.48. This variability is clinically important because it shows that FIB-4 remains useful as a simple first-line NIT, but fixed-score thresholding may not be sufficiently robust across external cohorts.

TabPFN achieved high specificity but lower sensitivity in the external cohorts. In Malaysia, TabPFN achieved a thresholded ROC-AUC of 0.69, specificity of 0.92, sensitivity of 0.45, and F1-score of 0.52. In India, it achieved a thresholded ROC-AUC of 0.66, specificity of 0.85, sensitivity of 0.47, and F1-score of 0.53. This pattern suggests that, under the fixed 0.50 threshold, TabPFN behaved conservatively, identifying fewer advanced fibrosis cases while maintaining a low false-positive rate. The probability-based ROC-AUCs were higher than the thresholded ROC-AUCs, reaching 0.81 in Malaysia and 0.69 in India, indicating that TabPFN retained useful continuous discrimination despite its conservative fixed-threshold operating profile.

The s-DNN achieved strong external performance in the Malaysian cohort and moderate performance in the Indian cohort. In Malaysia, it achieved a thresholded ROC-AUC of 0.77, specificity of 0.74, sensitivity of 0.81, and F1-score of 0.58, indicating a balanced sensitivity-specificity profile with high sensitivity. In India, it achieved a thresholded ROC-AUC of 0.67, specificity of 0.74, sensitivity of 0.59, and F1-score of 0.56. Thus, although performance remained cohort-dependent, the s-DNN retained a more balanced external operating profile across both external cohorts.

The LLM-based approaches showed moderate but inconsistent performance across cohorts. The zero-shot LLM achieved a ROC-AUC of 0.73 in the Malaysian cohort, with sensitivity of $0.81 \pm 0.02$, specificity of $0.65 \pm 0.02$, and F1-score of $0.52 \pm 0.02$. Fine-tuning improved the Malaysian ROC-AUC to 0.75 and F1-score to 0.56, but performance remained lower in the Indian cohort, with ROC-AUC values of 0.57 and 0.63 for zero-shot and fine-tuned settings, respectively.

\subsection{s-DNN calibration and explainability analysis}

\begin{figure}[!htbp]
\centering
\includegraphics[width=\textwidth]{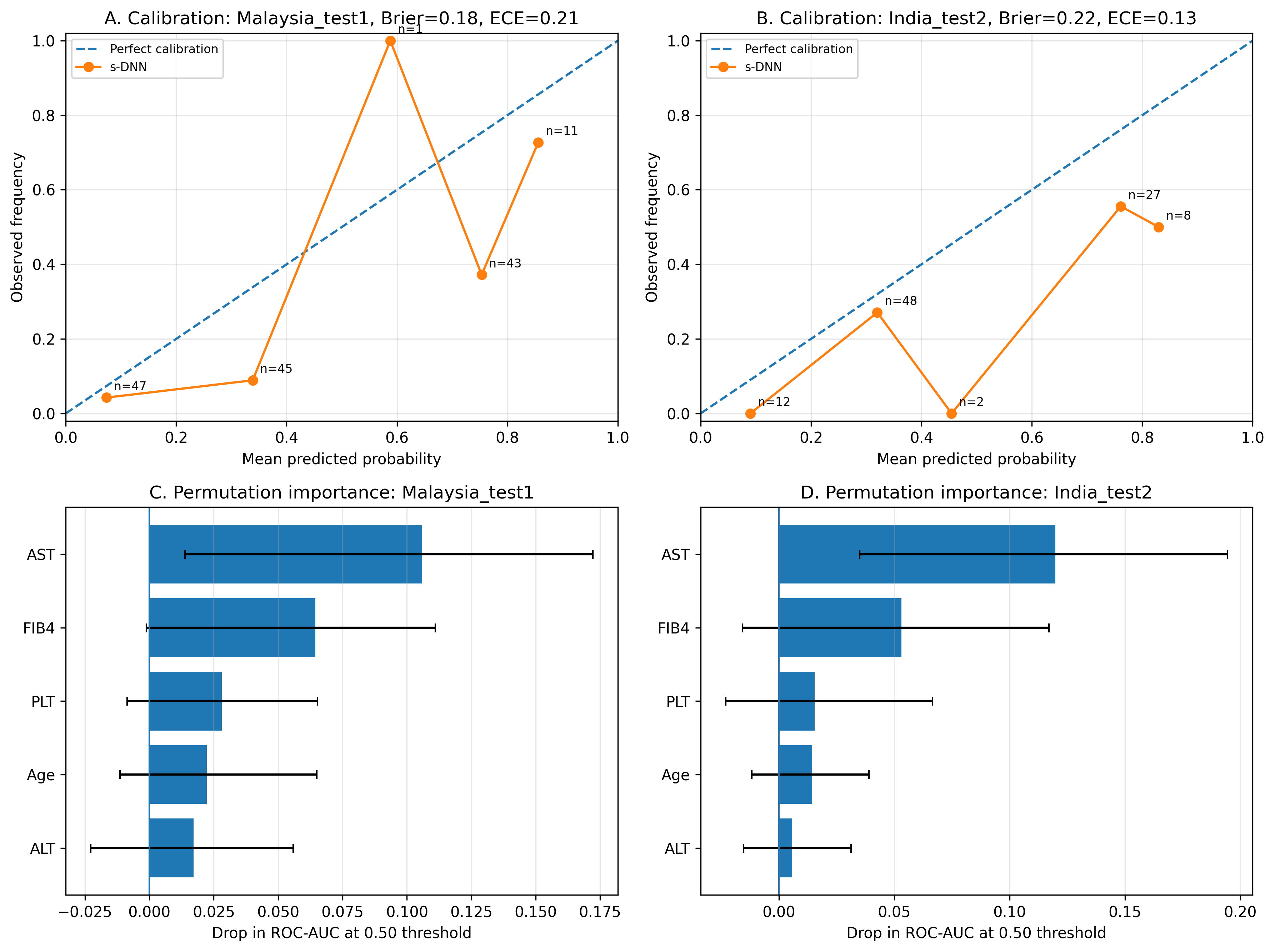}
\caption{External calibration and explainability analysis of the s-DNN MLE-NIT. Panels A--B show calibration curves for the Malaysian and Indian external validation cohorts, respectively. Points are annotated with the number of patients in each calibration bin. Panels C--D show permutation feature importance, measured as the decrease in ROC-AUC at the 0.50 classification threshold after feature permutation. Error bars represent variability across repeated permutations. AST and FIB-4 showed the largest contribution across both external cohorts.}
\label{fig:sdnn_external_calibration_xai}
\end{figure}

To further assess the external behavior of the s-DNN MLE-NIT, we performed calibration and explainability analyses on the Malaysian and Indian external validation cohorts (Figure~\ref{fig:sdnn_external_calibration_xai}). Calibration was assessed using reliability diagrams, Brier scores, and expected calibration error (ECE) \cite{steyerberg2010assessing,guo2017calibration}. Explainability was assessed using permutation feature importance, measured as the decrease in ROC-AUC at the 0.50 classification threshold after repeated random permutation of each input variable \cite{breiman2001random,altmann2010permutation}.

In the Malaysian cohort, the s-DNN achieved a Brier score of 0.18 and an ECE of 0.21. In the Indian cohort, the corresponding Brier score was 0.22 and the ECE was 0.13. These results suggest moderate calibration, although they should be interpreted cautiously because some calibration bins contained relatively few patients. The calibration analysis is therefore best viewed as an external diagnostic check rather than as definitive evidence of perfectly calibrated probabilities.

Permutation feature-importance analysis consistently identified AST and FIB-4 as the most influential variables across both external cohorts. In Malaysia, AST produced the largest decrease in ROC-AUC after permutation, followed by FIB-4. The same pattern was observed in India, where AST again showed the highest importance, followed by FIB-4. PLT, age, and ALT showed smaller and less stable contributions. This pattern supports the clinical plausibility of the s-DNN decision function, because the model's external predictions were primarily driven by variables already central to FIB-4-based fibrosis assessment.

\subsection{Clinical utility and interpretation of the FIB-4 variable space}
\label{sec:dca-results}

\begin{figure}[!htbp]
\centering
\includegraphics[width=\textwidth]{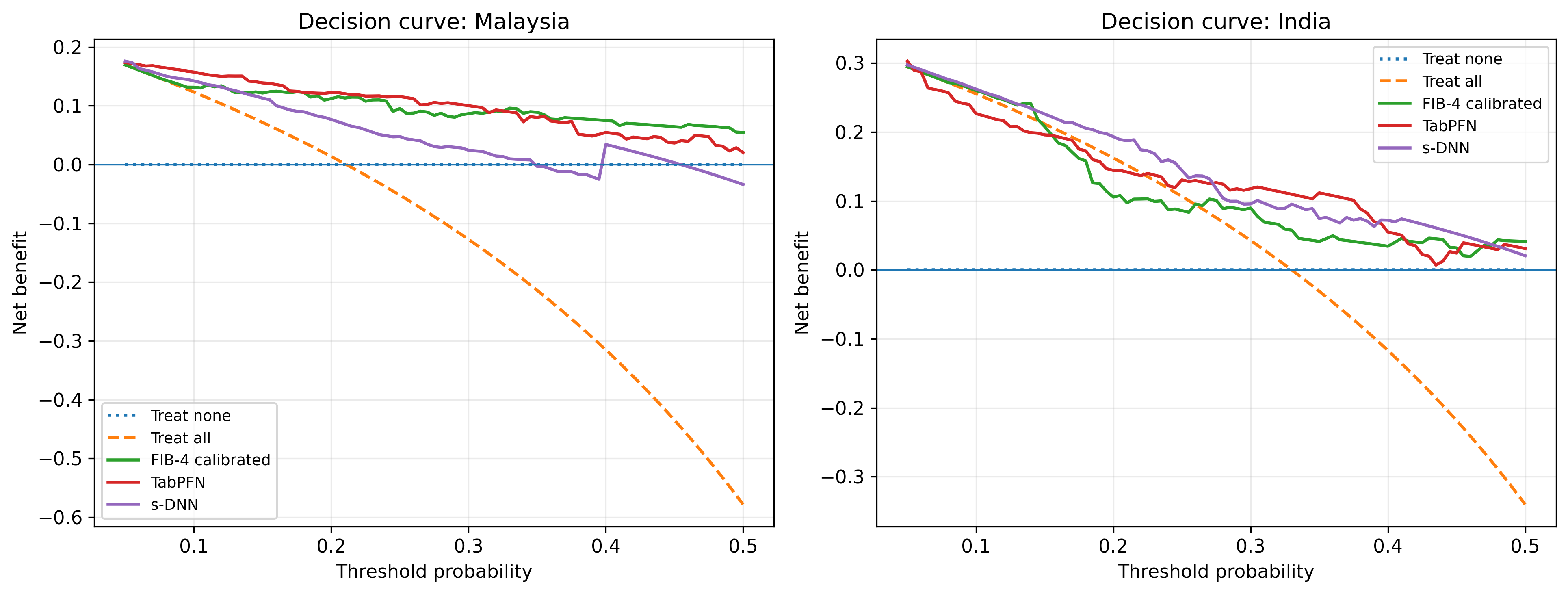}
\caption{Exploratory decision-curve analysis on the Malaysian and Indian external validation cohorts. Net benefit is shown across threshold probabilities from 0.05 to 0.50 for treat-none, treat-all, calibrated FIB-4, TabPFN, and s-DNN strategies. FIB-4 was represented by a logistic calibration model fitted on the Chinese training subset using FIB-4 as the sole predictor. TabPFN and s-DNN curves were generated from continuous model-predicted risks.}
\label{fig:dca_external}
\end{figure}

Decision-curve analysis showed cohort-dependent clinical utility patterns (Figure~\ref{fig:dca_external}). In the Malaysian cohort, TabPFN showed the highest average net benefit across the 0.05--0.20, 0.10--0.30, and 0.20--0.50 threshold-probability ranges, with mean net benefits of 0.15, 0.13, and 0.08, respectively. The calibrated FIB-4 risk model also performed strongly, with corresponding mean net benefits of 0.13, 0.11, and 0.08. The s-DNN showed lower average net benefit in Malaysia, particularly at higher threshold probabilities, with corresponding mean net benefits of 0.13, 0.08, and 0.02.

In the Indian cohort, the pattern differed. The s-DNN achieved the highest average net benefit across all examined threshold-probability ranges, with mean net benefits of 0.24, 0.19, and 0.09 across the 0.05--0.20, 0.10--0.30, and 0.20--0.50 ranges, respectively. TabPFN followed with mean net benefits of 0.22, 0.16, and 0.09, while the calibrated FIB-4 risk model showed lower average net benefit across the same ranges, with mean net benefits of 0.22, 0.15, and 0.06. Across the threshold grid, s-DNN was the best-performing model for 64.8\% of thresholds in India, whereas TabPFN was the best-performing model for 57.1\% of thresholds in Malaysia.

These findings suggest that the clinical utility of MLE-NIT models is cohort-dependent. While the s-DNN provided the most balanced threshold-based operating profile and the highest net benefit in the Indian cohort, TabPFN and calibrated FIB-4 showed stronger decision-curve performance in the Malaysian cohort. Therefore, DCA supports the use of MLE-NITs as clinically informative decision layers, but also highlights the need for local calibration, threshold selection, and prospective validation before deployment \cite{vickers2006decision}.

The central interpretation is that MLE-NIT should be viewed as a first, clinically conservative step into machine learning: it does not introduce new biomarkers or modify the care pathway, but instead applies a more flexible decision function to variables that clinicians already collect for FIB-4. This also explains why the performance gains are expected to be moderate rather than transformative. The input space remains deliberately limited to linearly and clinically associated routine variables; therefore, even non-linear models can improve the decision boundary only within the information limits of that restricted variable space.

\subsection{Foundation models did not uniformly outperform compact domain-specific learning}

The comparison between TabPFN, LLMs, and s-DNN suggests that foundation models are not automatically superior for low-dimensional structured clinical prediction. When evaluated on the external cohorts at the 0.50 classification threshold, TabPFN was more specificity-oriented, whereas the s-DNN provided a more balanced sensitivity-specificity profile, particularly in the Malaysian cohort. In addition, the s-DNN offered a transparent model-diagnostic layer through calibration and permutation-based explainability analyses, showing that its external predictions were mainly driven by AST and FIB-4.

This threshold-based evaluation is clinically important. For advanced fibrosis screening, sensitivity is important because false negatives may delay referral or further assessment, while specificity remains important to avoid unnecessary second-line testing. Therefore, model evaluation should report the operating-point performance used for clinical decision-making, including sensitivity, specificity, ROC-AUC at the selected threshold, and F1-score. Decision-curve analysis was additionally performed as an exploratory assessment of whether these operating-profile differences translated into net clinical benefit across risk thresholds; these analyses showed that the preferred model depended on the external cohort and threshold range.

These findings indicate that model suitability depends on the clinical task, the feature space, the decision threshold, and the intended sensitivity-specificity trade-off. LLMs can process structured prompts and may approximate clinical reasoning patterns, but this task depends on calibrated numerical relationships among laboratory variables rather than language understanding alone. Similarly, TabPFN is designed for tabular learning, but in this dataset its default 0.50-threshold predictions did not provide the balanced sensitivity-specificity trade-off required for screening-oriented advanced fibrosis detection.

\subsection{Positioning within the MLE-NIT research line}

The present study extends a coherent line of previous work on machine-learning-enhanced fibrosis and steatosis assessment in liver disease. The earliest part of this line used compact DNNs with structured ultrasound-derived and demographic variables for fibrosis and NAFLD/steatosis classification \cite{angelakis2018aium_cld_bmode_dnn,angelakis2018ecr_fibrosis_bmode_hemodynamic_dnn,angelakis2018acr_nafld_bmode_dnn}. Angelakis and Chen then showed that CatBoost with feature engineering could outperform FIB-4 using FIB-4-related variables on the same China, Malaysia, and India cohorts used in the present study \cite{angelakis2023fib4parameters}. A later study demonstrated that CTGAN-based synthetic data generation could further improve a CatBoost model trained on the same FIB-4 variable family \cite{angelakis2024catboost}. In parallel, the MLE-NIT framework was introduced to combine multiple non-invasive tests and their underlying variables using compact DNNs \cite{angelakis2025sdnnnits}.

The current study adds a new contribution to this sequence. Instead of asking whether machine learning can outperform FIB-4, which prior work had already suggested, we asked whether current state-of-the-art model families are necessary or advantageous for this task under the same restricted FIB-4 variable space. The answer appears to be nuanced. General-purpose models such as LLMs and tabular foundation models can be applied, but their external operating profiles were not uniformly balanced. Therefore, the most promising direction for enhanced non-invasive fibrosis assessment may not be the largest available model, but the best clinically constrained non-linear model.

\subsection{Clinical and methodological implications}

The main implication is that FIB-4 should be considered a baseline NIT, not the ceiling of what can be achieved with its variables. The same variables that make FIB-4 practical can also support MLE-NIT models that preserve clinical simplicity while improving diagnostic flexibility. This is especially important in MASLD, where patients are heterogeneous and fixed thresholds may not capture population-specific risk patterns.

Methodologically, the results caution against assuming that foundation models are automatically optimal for biomedical tabular prediction. For structured clinical tasks with few variables and limited sample sizes, compact architectures may provide a favorable bias-variance and deployment trade-off. The s-DNN used here is small, task-specific, and directly optimized for the fibrosis endpoint. Its architecture follows a simple design rule: two or three hidden layers, with an odd prime number of nodes per hidden layer. In the present instantiation, the s-DNN contained only 354 trainable parameters and required 0.0045 MB as a serialized PyTorch \texttt{state\_dict}. By contrast, TabPFN contained 7,244,554 parameters and required 28.0648 MB as a serialized fitted classifier object. This corresponds to approximately four orders of magnitude fewer parameters and more than three orders of magnitude smaller serialized size for the s-DNN. Therefore, the s-DNN is not only a clinically constrained MLE-NIT, but also a substantially more lightweight model for potential deployment.

The present results should not be interpreted as evidence that the s-DNN discovered new fibrosis biology. Rather, the model appears to re-express known FIB-4-related risk gradients through a compact non-linear decision layer. This is consistent with the MLE-NIT design goal: to improve the use of existing clinical variables without adding workflow complexity. The modest magnitude of improvement is therefore expected, because the information content is intentionally limited to the same routine variables that define the original FIB-4 pathway.

The calibration and explainability analyses further strengthen the clinical interpretation of the s-DNN. Calibration diagnostics provide an initial assessment of whether model probabilities behave consistently across external cohorts \cite{steyerberg2010assessing}, while permutation importance offers a model-agnostic explanation of which variables most strongly influence the operating-point performance \cite{breiman2001random,altmann2010permutation}. The observation that AST and FIB-4 were the dominant contributors is clinically plausible and supports the interpretation that the s-DNN learned within the intended FIB-4 variable space rather than relying on unstable or unrelated signals.

\subsection{Limitations}

This study has a few limitations. First, the feature set was intentionally restricted to variables from the FIB-4 variable space. This improves clinical simplicity, preserves workflow invariance, and isolates the effect of modelling, but it also limits the maximum achievable performance compared with models incorporating additional laboratory markers, imaging, elastography, omics features, or clinical text. The observed performance gains should therefore be interpreted as improvements within a deliberately constrained clinical feature space, rather than as evidence that FIB-4 variables alone can fully capture the biological complexity of advanced fibrosis.

Second, the Chinese cohort was split into a training subset and an internal validation/tuning subset; the latter was used only for model selection or tuning where applicable and is not reported as an independent result. The 0.50 classification threshold used for the machine-learning models was fixed a priori as the operating point for external model comparison \cite{powers2011evaluation}. It was not selected or optimized using the Chinese internal validation/tuning subset, nor was it optimized on the Malaysian or Indian external validation cohorts. An exploratory threshold audit on the Chinese internal validation/tuning subset suggested alternative thresholds of 0.52 for TabPFN and 0.71 for s-DNN, but these thresholds were not used to generate the reported external validation metrics. Future studies should prospectively define clinically meaningful operating thresholds according to the intended use case, such as rule-out screening, referral prioritization, or rule-in confirmation.

Third, the external validation cohorts were modest in size and differed in prevalence and clinical characteristics, which may explain part of the observed performance variability. Although bootstrap confidence intervals were estimated for external ROC-AUCs using 2,000 resamples \cite{efron1993bootstrap,carpenter2000bootstrap}, the study was not powered for definitive pairwise superiority testing between models. Formal statistical comparisons, including tests such as DeLong's test for correlated ROC curves and confidence intervals for model differences, should be added in larger validation studies before claiming definitive superiority.

Fourth, calibration and explainability analyses were performed only on the Malaysian and Indian external validation cohorts. Some calibration bins contained few patients, and permutation-importance estimates may be unstable in small samples \cite{breiman2001random,altmann2010permutation}. These analyses should therefore be interpreted as exploratory model diagnostics rather than definitive evidence of probability calibration or mechanistic explanation. In addition, the permutation-importance analysis identifies the contribution of variables to model performance, but it does not establish causal relationships or new fibrosis biology.

Fifth, decision-curve analysis was exploratory \cite{vickers2006decision}. FIB-4 was represented in DCA as a calibrated single-predictor risk model rather than as the original threshold-only clinical rule, because net-benefit estimation requires continuous risk scores. TabPFN and s-DNN were evaluated using continuous model-predicted risks, whereas LLM-based approaches were excluded from the main DCA because only classification outputs were available. The cohort-dependent DCA findings should therefore be interpreted as hypothesis-generating and should be reassessed in prospective clinical workflow studies with prespecified decision thresholds and clinical action pathways.

Sixth, LLM performance may depend on prompt design, fine-tuning settings, output parsing, and model version, although the use of a fixed dated snapshot reduces model-drift concerns. Because only classification outputs were available for the LLM workflows, LLMs could not be fully assessed in probability-based analyses such as calibration and DCA. Future work should evaluate whether calibrated LLM-derived risk scores can be obtained reliably and whether they add value beyond conventional tabular models.

Seventh, the odd-prime width rule used in the s-DNN is a transparent compact-architecture heuristic intended to constrain model complexity in small structured clinical datasets. This study did not perform an ablation against non-prime compact architectures, alternative layer widths, or broader neural architecture search. Therefore, the present results support the feasibility of this compact s-DNN design, but not the universal optimality of the odd-prime width rule.

Finally, while the s-DNN demonstrates promising external performance and clinically plausible feature-importance patterns, real-world deployment requires attention to integration workflows, clinician--AI interaction design, calibration monitoring, and monitoring for dataset shift. Future work should evaluate prospective clinical utility, including impact on referral decisions and patient outcomes, as well as strategies for continuous model updating in evolving healthcare settings.

Despite these limitations, the study provides a clinically relevant comparison: under the same simple FIB-4-derived feature constraints, a compact s-DNN provided a balanced external operating profile and clinically plausible model diagnostics without requiring additional biomarkers, imaging, omics data, or changes to the existing clinical workflow.

\section{Conclusion}
\label{sec:conclusion}

This study evaluated whether modern artificial intelligence models can enhance advanced fibrosis detection in MASLD while using only routinely available variables from the FIB-4 variable space. Across biopsy-confirmed cohorts from China, Malaysia, and India, FIB-4 showed variable external performance, confirming the limitations of fixed-score thresholding. General-purpose foundation models, including TabPFN and LLM-based approaches, were applicable, but their thresholded operating profiles were not uniformly balanced across external cohorts.

The s-DNN achieved the most balanced fixed-threshold operating profile among the evaluated models, with strong sensitivity and thresholded ROC-AUC in the Malaysian cohort and moderate external performance in the Indian cohort. TabPFN achieved high specificity at the 0.50 threshold, but its predictions were more conservative, with lower sensitivity. Probability-based ROC-AUC analysis provided a complementary view: calibrated FIB-4 showed the highest continuous discrimination in the Malaysian cohort, whereas s-DNN and TabPFN showed stronger probability-based discrimination than FIB-4 in the Indian cohort. These results indicate that model ranking depends on the evaluation perspective: fixed-threshold clinical operating performance, continuous discrimination, calibration, or net clinical benefit.

External calibration and permutation-based explainability analyses further suggested that the s-DNN behaved plausibly, with AST and FIB-4 emerging as the dominant contributors across external cohorts. The brief feature-ablation analysis in the Chinese training subset suggested that the FIB-4 variable space is partly redundant, and that the s-DNN distributes information across correlated clinical inputs rather than relying exclusively on a single predictor. Exploratory decision-curve analysis added a more nuanced clinical-utility perspective: the s-DNN achieved the highest average net benefit across the evaluated threshold ranges in the Indian cohort, whereas TabPFN and calibrated FIB-4 showed stronger decision-curve performance in the Malaysian cohort.

These findings support the view that the future of non-invasive fibrosis assessment is not necessarily to replace simple clinical scores with larger models, additional biomarkers, or more expensive data modalities, but to enhance clinically established NITs through carefully designed non-linear machine learning. In this setting, s-DNN-based MLE-NITs offer a practical and scientifically grounded pathway beyond conventional FIB-4: the clinical input remains the same, but the decision function becomes more adaptive, data-driven, externally diagnosable, and capable of modelling non-linear risk patterns within the constraints of the original feature space. The model-complexity analysis further supports this interpretation: the s-DNN achieved its external operating profile with only 354 trainable parameters, compared with more than 7.2 million parameters for TabPFN, highlighting the potential value of small, task-specific architectures for workflow-preserving clinical machine learning. At the same time, the cohort-dependent probability-based and decision-curve findings show that local calibration, threshold selection, and prospective workflow validation remain necessary before clinical deployment.

\bibliographystyle{unsrtnat}
\bibliography{biblio}

\end{document}